\documentclass[10pt,twocolumn,letterpaper]{article}

\usepackage{cvpr}
\usepackage{times}
\usepackage{epsfig}
\usepackage{graphicx}
\usepackage{amsmath}
\usepackage{amssymb}
\usepackage{authblk}
\usepackage{bm}

\usepackage[pagebackref=true,breaklinks=true,letterpaper=true,colorlinks,bookmarks=false]{hyperref}

\cvprfinalcopy 

\begin{document}

\title{Re-rank Coarse Classification with Local Region Enhanced Features for Fine-Grained Image Recognition}




\author{
	Shaokang Yang\thanks{Corresponding author (equal contribution)},
	\;\; Shuai Liu\thanks{Corresponding author (equal contribution)},
	\;\; Cheng Yang,
	\;\; Changhu Wang\\	
	ByteDance AI Lab, Beijing, China\\	
	{\tt\small \{yangsk0205,sliu012\}@gmail.com, \{yangcheng.iron, wangchanghu\}@bytedance.com}
}


\maketitle

\begin{abstract}
   Fine-grained image recognition is very challenging due to the difficulty of capturing both semantic global features and discriminative local features. Meanwhile, these two features are not easy to be integrated, which are even conflicting when used simultaneously. In this paper, a retrieval-based coarse-to-fine framework is proposed, where we re-rank the TopN classification results by using the local region enhanced embedding features to improve the Top1 accuracy (based on the observation that the correct category usually resides in TopN results). To obtain the discriminative regions for distinguishing the fine-grained images, we introduce a weakly-supervised method to train a box generating branch with only image-level labels. In addition, to learn more effective semantic global features, we design a multi-level loss over an automatically constructed hierarchical category structure. Experimental results show that our method achieves state-of-the-art performance on three benchmarks: CUB-200-2011, Stanford Cars and FGVC Aircraft. Also, visualizations and analysis are provided for better understanding.
\end{abstract}

\section{Introduction}
\begin{figure}[t]
\begin{center}
\includegraphics[width=1.0\linewidth]{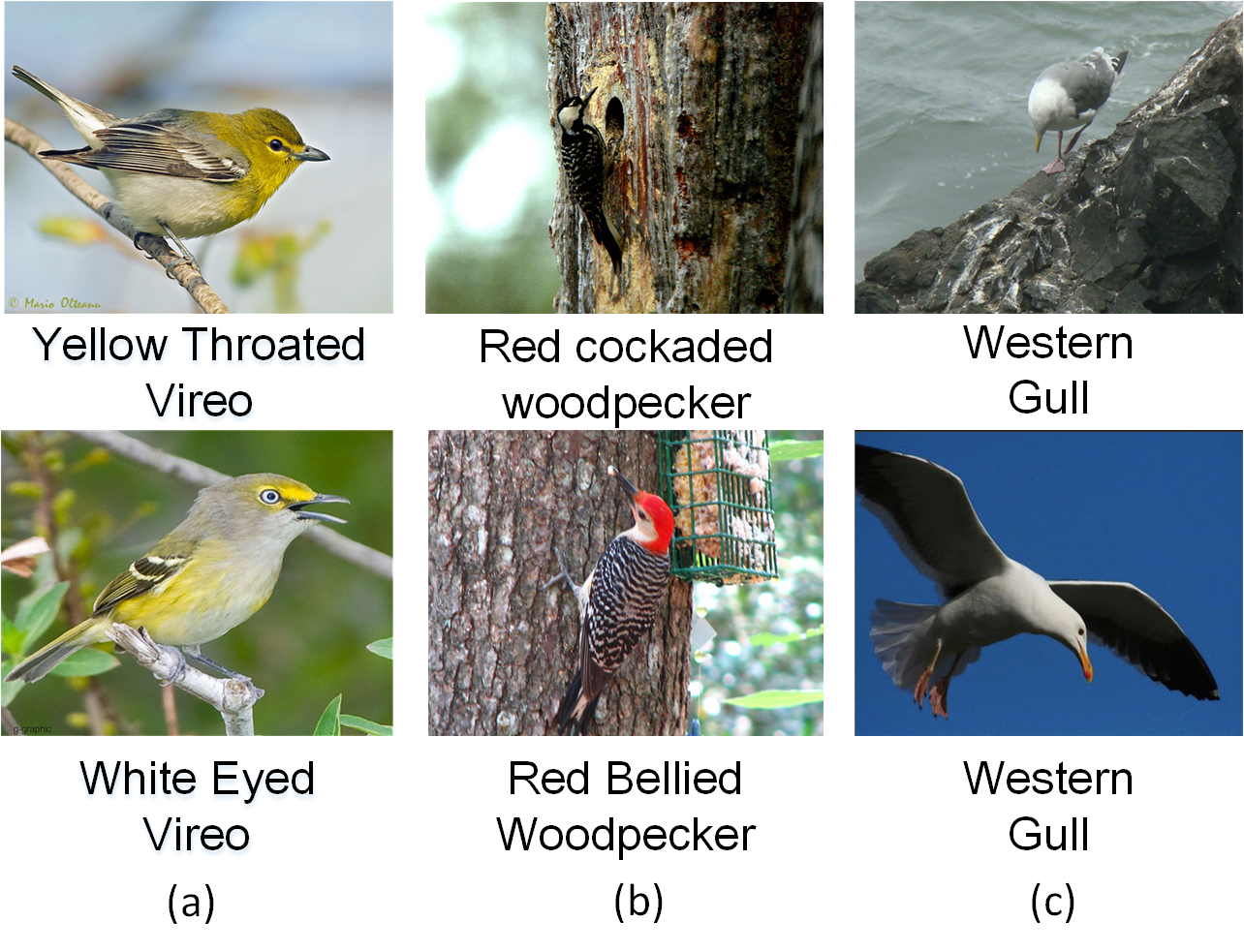}
\end{center}
   \caption{Some hard instances from the CUB-200-2011\cite{wah2011caltech}. Sub-figure (a) and (b) show very similar categories, where the subtle differences lie in the color of certain parts. Sub-figure (c) shows the different poses of the same category.}
\label{fig1}
\end{figure}
Fine-grained visual classification (FGVC) aims at recognizing subordinate categories that belong to the same superior class, such as distinguishing different kinds of wild birds\cite{wah2011caltech}, cars\cite{krause20133d} or airs\cite{maji2013fine}. Unlike general classification tasks, fine-grained sub-categories usually have similar appearances, which can only be classified by some subtle details. For instance, the difference between \textit{Yellow Throated Vireo} and \textit{White Eyed Vireo} is the color of certain parts as shown in Fig~\ref{fig1} (a). 

To address this task, lots of methods were proposed that can be divided into global-based methods\cite{li2017dynamic,cui2018large, SimonRDD20} and part/attention-based methods\cite{YangLWHGW18,GeLY19,korsch2019classification,du2020fine,abs-2004-02684,ZhuangW020}. 
Most of them aim at learning more effective global/part features separately and adopt a softmax to predict final classification results. Unlike these methods, in this work, we propose a novel retrieval-based model that can simultaneously learn effective global features and discriminative local features inspired by DELG\cite{DBLP:journals/corr/abs-2001-05027}. We refer to the new model as Coarse Classification and Fine Re-ranking (CCFR), where local region enhanced features are generated to re-rank the classification results predicted by the conventional global pooling features. Specifically, our model includes two main branches. One targets to select $topn$ classification results (named as Coarse Classification) predicted by global features, the other aims at learning discriminative local-region features that are fused with global features to build a searching database for re-ranking the Coarse Classification results. By this way, our CCFR can make use of the local region enhanced features to correct images which cannot be distinguished accurately only relying on global features. For instance, merely depending on global features failed when the object scale is too small in a full image or the subtle difference lies in some local regions as shown in Fig~\ref{fig1}. 

Therefore, the key to our model is how to accurately localize the discriminative part regions. Recently, a number of methods\cite{Berg2013POOFPO,BransonHBP14,chai2013symbiotic,gavves2013fine,liu2012dog,xie2013hierarchical,ZhangDGD14} that utilized the object/part annotations (e.g. bird parts annotation in bird fine-grained classification) were proposed to accurately localize part regions. However, obtaining object/part annotations is very expensive.
In this work, we propose a weakly-supervised method to effectively localize discriminative part regions with only image-level labels. Specifically, we leverage a Feature Pyramid Network (FPN)\cite{lin2017feature} to generate a series of multi-scale regions, where triplet loss\cite{schroff2015facenet} is used to make local regions more discriminative other than informative with ranking loss\cite{YangLWHGW18}. It is reasonable that discriminative local regions (e.g., bird's head, bird's wings) are more important for the FGVC task than the whole object regions containing the highest informativeness. Visualizations are provided in section 4.5. In addition, to learn more effective global semantic features, we utilize an unsupervised method to automatically construct a hierarchical category structure and then design a Multi-level loss to train our model. 
Overall, the main contributions of our work are listed as follows:
\begin{itemize}
    \item We build a Coarse Classification and Fine Re-ranking(CCFR) architecture that simultaneously uses both global and part features by re-ranking.
    \item We propose a weakly-supervised method to effectively select discriminative local regions without object/part annotations.
    \item We utilize an unsupervised method to automatically construct a hierarchical category structure to learn more effective global semantic features.
\end{itemize}

\section{Related Works}
Fine-grained image classification aims at recognizing the objects of the sub-categories from visually similar category and has been studied for many years. Existing methods can be roughly categorized into two types: global-based methods\cite{li2017dynamic,cui2018large, SimonRDD20} and part/attention-based methods\cite{LinSLJ15,ZhangSGD15,YangLWHGW18,GeLY19,korsch2019classification,du2020fine,abs-2004-02684,ZhuangW020}. The former focuses on how to extract more effective pre-training parameters and design more useful pooling techniques. The latter targets at training a detection network for localizing part regions which are used to perform classification.

\subsection{Global-based methods for FGVC}
Cui et al.\cite{cui2018large} utilizes large-scale datasets (e.g., ImageNet\cite{deng2009imagenet} and Inaturalist\cite{van2015building}) to learn more effective pre-training parameters for domain-specific FGVC tasks. To resolve the difference between these pre-training datasets and the target fine-grained datasets, they propose a measure to estimate domain similarity via Earth Mover’s Distance. And then they pre-selects certain classes from the large-scale datasets which match best to the current fine-grained datasets. Similarly, Krause et al.\cite{KrauseSHZTDPF16} collect images from the Internet to enrich the FGVC datasets proving the effectiveness of large-scale datasets for FGVC performance.
Other methods aim at developing advanced pooling strategies. For instance, Lin et al.\cite{LinRM15} propose a bilinear structure to compute the pairwise feature interactions by two independent CNN. Similar works are \cite{SimonGDDR17,SimonRDD20,ZhengFZL19}. Besides, researchers also combine Fisher vectors to enhance the global representations \cite{abs-2007-02080,ZhangXZLT16}. All of these works target at enriching the global features representation to obtain better classification results. 
In our model, to learn more effective global semantic features, we propose a Multi-level loss which depends on an automatically constructed hierarchical category structure.
\subsection{Part-based methods for FGVC}
One can design a part-based classification method by employing the part annotations if they exist. However, such annotations are expensive and usually unavailable, and the current researchers mainly focus on localizing part-region with only image-level labels\cite{BencyKLKM16,KimCY17,SinghL17,OquabBLS15,ZhouKLOT16,zhangWF0H18,WangWYLLL20}. He et al.\cite{he2019and} propose a sophisticated reinforcement learning method to estimate how many and which image regions are helpful to distinguish the categories. Ge et al.\cite{GeLY19} use Mask R-CNN and CRF-based segmentation alternately to extract rough object instances. However, such a multi-stage training process is hard to implement which significantly hamper the availability in practical use. Yang et al.\cite{YangLWHGW18} adopt a ranking loss to train an FPN network to obtain the most informative regions. However, such a loss function encourages the model to select boxes containing entire objects other than discriminative local regions. To overcome such a problem, we utilize the triplet loss to select more discriminative regions.

Cao et al.\cite{DBLP:journals/corr/abs-2001-05027} design a DELG model that unifies global and local features into a single deep model to obtain accurate retrieval results on Image Retrieval Tasks. 
Inspired by this work, in our method we design a re-ranking policy to improve the performance by correcting the uncertain $topn$ classification results.
\begin{figure*}
\begin{center}
\includegraphics[width=1.0\linewidth]{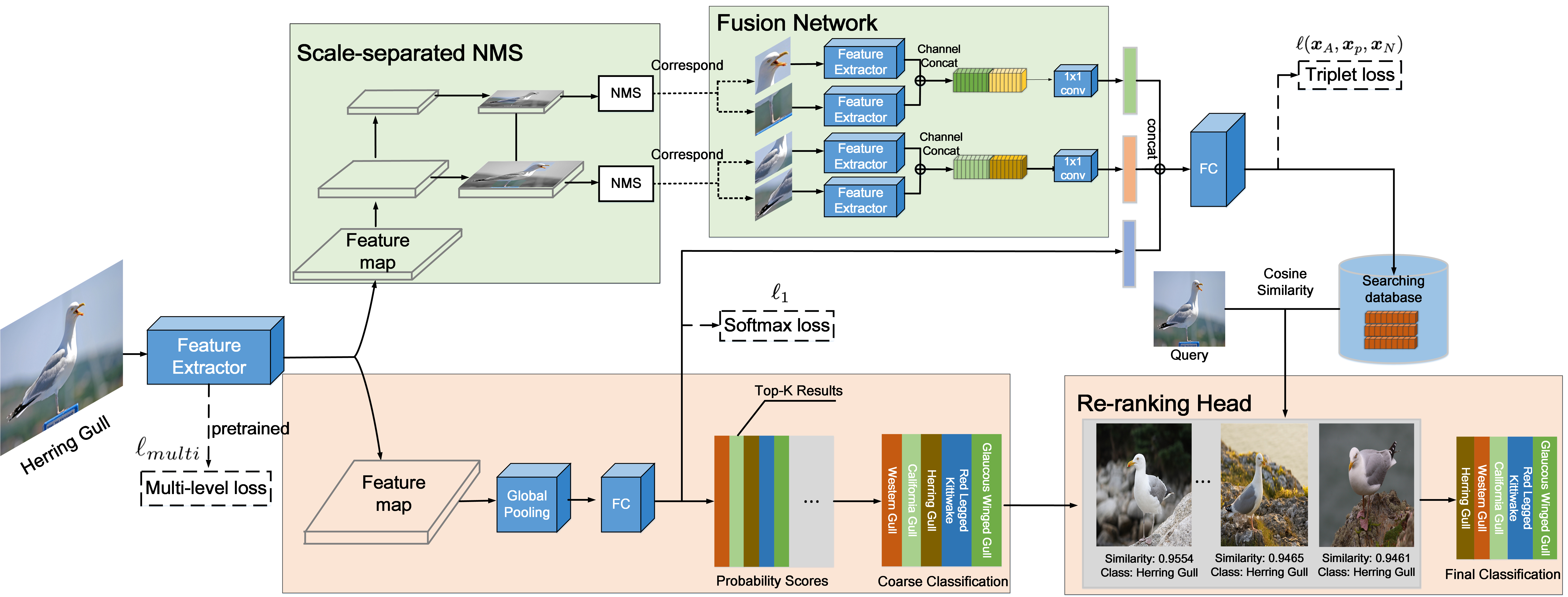}
\end{center}
   \caption{The overview of our CCFR architecture. It mainly includes two branches: 1) On the top branch, we utilize the Triplet loss and the Scale-separated NMS to discover more discriminative local regions, and integrate these region features with the whole image features to obtain the final embedding features, which are used to build a searching database. 2) On the bottom branch, we first obtain the Coarse Classifications (the $topn$ Softmax probabilities), and then re-rank them by the statistic of the retrievals from the searching database. The Feature Extractor is pretrained by a proposed Multi-level loss and is updated by the combination of a Triplet loss and a Softmax loss during training.}
\label{fig2}
\end{figure*}
\section{Method}
In this section, we describe our Coarse Classification and Fine Re-ranking (CCFR) architecture in detail. The overview is shown in Fig~\ref{fig2}, which includes two main branches: 1) The top branch first extracts discriminative local-region features used to enhance global features, and then builds a database for re-ranking. 2) The bottom branch first acquires the $topn$ classification results with global features, and then re-ranks them with the constructed searching database to obtain the final results. As shown in\cite{li2017dynamic,cui2018large,SimonRDD20}, satisfactory results were obtained based on global features, however, these methods usually fail in some cases depending on subtle parts (e.g., Fig~\ref{fig1}). To solve such problems, a series of methods\cite{YangLWHGW18,GeLY19,korsch2019classification,du2020fine,abs-2004-02684,ZhuangW020} have been proposed to learn local features. However most of them only adopt the convention softmax classification.
Unlike these methods, our CCFR is a retrieval-based method which can effectively correct the misclassified images with the combination of the $topn$ classification and searching results.
Besides, comparing with the current state-of-the-art StackcLSTM\cite{GeLY19} which is a very messy multi-stage training model, the model in our method is easier to implement and train.

\subsection{Multi-level loss for learning effective global features}

\begin{figure}[t]
\begin{center}
\includegraphics[width=1.0\linewidth]{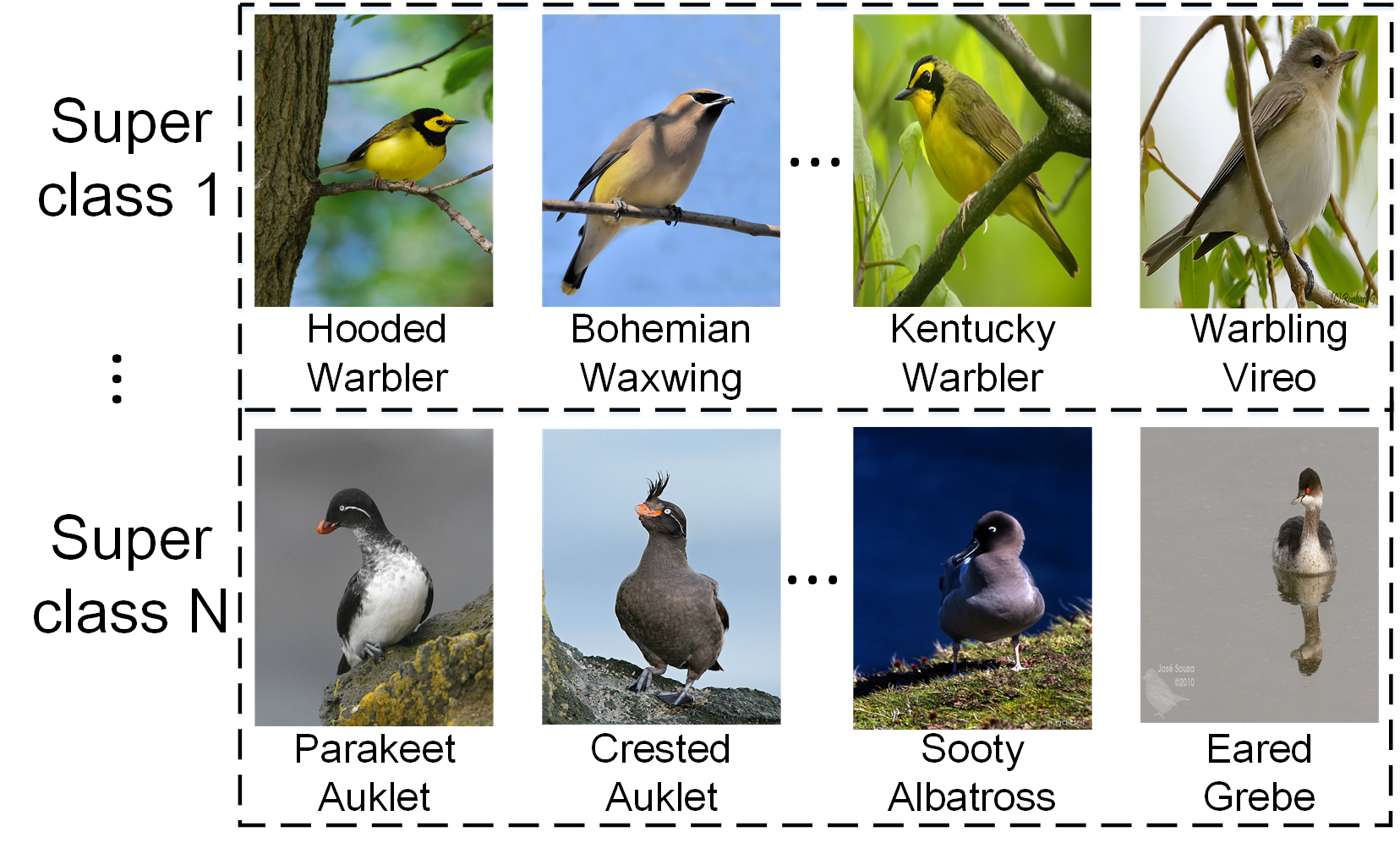}
\end{center}
   \caption{The generated hierarchical categories by clustering. Birds with similar appearances are grouped into the same super class. }
\label{superclass}
\end{figure}

As we have known, feature expression is fundamental for visual classification tasks. How to learn effective features is the key to these tasks. 
The human visual cognition system processes from rough to fine-grained. (e.g., we first realize that it belongs to a bird at a glance, and then we confirm that it belongs to \textit{Vireo} or other bird types through careful observation.) 
To imitate the above cognition process, we organize the categories into Multi-level loss according to the visual appearance.
Specifically, we first automatically construct a hierarchical category system through visual feature clustering. As shown in the Fig~\ref{superclass}, the four birds on the top (bottom) are more similar to each other which can be naturally grouped into one super class. We then propose a Multi-level loss to utilize the constructed category system. We adopt two standard Softmax loss on the children (origin) category and super (cluster) category and then design a constraint loss between these two categories:
\begin{eqnarray}
    \ell_1 &=& -\sum_{i=1}^C\bm{y_i}\log \bm{p_i} \\
    \ell_2 &=& -\sum_{j=1}^{C_f}\bm{y_j}\log\bm{p_j}\\
    \ell_h &=& max(0, \bm{p_{children}} - \bm{p_{parent}})
\end{eqnarray}
where $\bm{p_i}$ is the probability of class $i$ in children categories, $\bm{p_j}$ is the probability of class $j$ in super categories, $\bm{p_{children}}$ denotes the average of all children category probabilities belonging to the same super category, and $\bm{p_{parent}}$ is the corresponding super category probability, $C$ is the number of children classes, $C_f$ is the number of super classes, and $\ell_1$ and $\ell_2$ are the standard Softmax loss. $\ell_h$ is the constraint loss between super category probability and children's. 
The final Multi-level loss is defined as follows:
\begin{equation}
    \ell_{multi} = \ell_1 + \lambda * \ell_2 + \ell_h
\end{equation}
where $\lambda$ is the hyper-parameter. In our setting, $\lambda = 1$.

\subsection{Localizing discriminative local regions with weakly supervising}
To select discriminative local regions, we leverage a top-down pathway with lateral connections detection network inspired by the Feature Pyramid Networks (FPN)\cite{lin2017feature} as shown in Fig~\ref{fig2}. We can obtain a series of feature maps with different spatial resolutions. By using these multi-scale feature maps, we can generate local regions among different scales and ratios. Specifically, for an input image of size 448, we set the bounding-box scales to \{96, 192\} and aspect ratio to 1:1. Given that different parts of the object own different scales (e.g., the bird's head area is smaller than its wings), performing NMS over boxes of all scales may result in larger boxes suppressing the smaller ones.
Thus, we perform NMS on each scale separately (named as Scale-separated NMS). Then we resize the selected boxes to the pre-defined size (e.g., 224$\times$224) and feed them into the feature extractor to obtain local features. To prevent overfitting, we share parameters of feature extractor used in local regions with the one used in global features. To better capture the spatial relationships among the different regions, for each scale, we concatenate the local features along the channel direction and fuse them with $\bm{1\times1}$ convolution filters as shown in Fig~\ref{fig2} (we name it as Fusion Network).

\begin{figure*}[t]
\begin{center}
\includegraphics[width=0.9\linewidth]{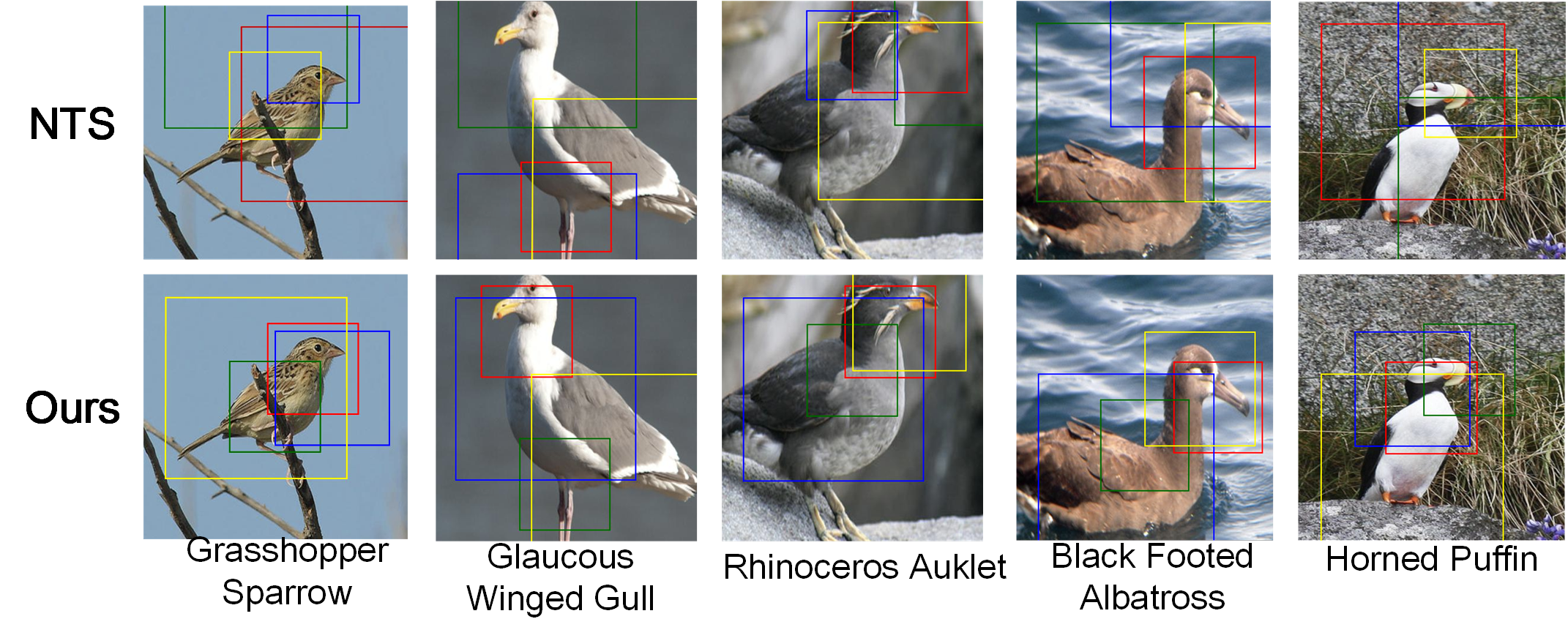}
\end{center}
   \caption{The Top4 selected boxes by NTS and our CCFR. Compared to NTS (the top row), our model tends to focus on smaller boxes containing discriminative parts.}
\label{nts2ours}
\end{figure*}

The key of this process is how to select discriminative local regions.
Given that the part/object annotations for FGVC are expensive, it is desirable to train the model with only image-level labels (weakly supervised). 
As mentioned above, the NTS utilized a ranking loss to select informative regions, other than discriminative regions, leading to a preference for larger boxes (as shown in the top row of Fig~\ref{nts2ours}). 
Triplet loss\cite{schroff2015facenet} has shown beneficial for learning
discriminative local features by maximizing distance of different classes and minimizing the distance of the same class.
Thus, we adopt the triplet loss to learn discriminative local features, which can be defined as follows:
\begin{equation}
\begin{split}
        \ell(\bm{x}_A, \bm{x}_p, \bm{x}_N) & =   \bm{f}(sim(\bm{x}_A,\bm{x}_P) \\
        & -  sim(\bm{x}_A, \bm{x}_N) - \bm{a})
\end{split}
\end{equation}
where $sim$(·) denotes the cosine similarity, $\bm{x}_A$ is an anchor input, $\bm{x}_P$ denotes a positive input of the same class as $\bm{x}_A$, $\bm{x}_N$ is a negative input of a different class from $\bm{x}_A$, $\bm{a}$ is a margin between positive and negative pairs, $\bm{f}$($\cdot$) is defined as follows:
\begin{equation}
    \bm{f}(y) = max(0, -y)
\end{equation}
where $y$ is the value of $sim(\bm{x}_A,\bm{x}_P) - sim(\bm{x}_A,\bm{x}_N) - \bm{a}$.
Each of ($\bm{x}_A$), ($\bm{x}_P$) and ($\bm{x}_N$) is the embedding features by concatenating the fused local features and global features.
Finally, the total loss is defined as:
\begin{equation}
    \ell =   -\sum_{i=1}^C \bm{y_i}\log \bm{p_i}  
     + \mu*\ell(\bm{x}_A^i,\bm{x}_P^i,\bm{x}_N^i)
\end{equation}
where $\mu$ is the hyper-parameter, in our setting $\mu = 1$.

\subsection{Re-rank the coarse classification by retrieval}

Up to now, we have obtained the $topn$ probability scores (coarse classification results) and discriminative local enhanced features (anchor $\bm{x}_A$ features). Then we construct the searching database using these features extracted from the training samples. We treat the image that needs to be reorganized as a query ($\bm{q}$), and by computing the cosine similarity between the query and the searching database we obtain the $topm$ similarity scores with their labels. We finally re-rank the classification results based on these similarity scores and labels. 

The detailed re-ranking formulation is listed as follows.
Let $\bm{X}$=\{$\bm{x}_k^{c}$\} be the selected training samples as the searching database, where $\bm{x}_k^{c}$ is the local enhanced feature vector influenced by our model ($k=1,2,...,N$ is the sample index, $c=1,2,...,C$ is the class label index).
Again, let $sim(\bm{q},\bm{X})$=\{$sim(\bm{q},\bm{x}_k^{c})$\} be the similarity of the testing sample $\bm{q}$ between all the training samples in the database. $rank(\bm{q},\bm{X}, topm$) is the most similar $topm$ training samples of testing $\bm{q}$ measured by $sim(\bm{q},\bm{X})$

Then the finally re-ranked score of the testing sample $\bm{q}$ w.r.t class ${c}$ is

\begin{equation}
{S_{\bm{q}}^c =
\left\{
\begin{array}{rl}
S_f(\bm{q},c),  {S_f(\bm{q},c)  >= T_{sf}} \\
\alpha*S_f(\bm{q},c)+\beta*Sc(\bm{q},c),{S_f(\bm{q},c) < T_{sf}}\label{con:eq8}
\end{array}
\right.}
\end{equation}
where $\alpha$ and $\beta$ are the weights balancing the two terms, $S_f(\bm{q},c)$ is the Softmax probability of $\bm{q}$ w.r.t class $c$, $T_{sf}$ is a threshold for softmax score, and $Sc(\bm{q},c)$ is the normlized similarity score between query and the class $c$ which is calculated as 
\begin{equation}
\begin{aligned}
{Sc(\bm{q},c)}=\sum_{k=1}^{N_c}sim(\bm{q},\bm{x}_k^c)/\sum_{c=1}^{topn}\sum_{k=1}^{N_c}sim(\bm{q},\bm{x}_k^c) \\
s.t. sim(\bm{q},\bm{x}_k^{c})>T_{sc}, \bm{x}_k^{c} \in rank(\bm{q},\bm{X},{topm})
\end{aligned}
\end{equation}
where $T_{sc}$ is a threshold score for searching similarity.
In this way, we could probably distinguish and correct the extremely similar sub-categories which usually reside in the $topn$ classification results with non-dominant softmax scores.

\begin{table}
\begin{center}
\begin{tabular}{|l|c|c|c|}
\hline
Datesets & Train & Test & Category \\
\hline\hline
CUB-200-2011 & 5,994 & 5,794 & 200 \\
\hline
FGVC Aircraft & 6,667 & 3,333 & 100 \\
\hline
Stanford Cars & 8,144 & 8,041 & 196 \\
\hline
\end{tabular}
\end{center}
\caption{Statistics of three datasets.}
\end{table}

\begin{table*}
\begin{center}
\begin{tabular}{|l|c|c|c|c|c|}
\hline
Method & Base Model  & CUB Acc.(\%) & Airs Acc.(\%) &  Cars Acc.(\%) \\
\hline
\hline
ResNet-50\cite{li2017dynamic}  & ResNet-50  & 84.5 & - & - \\
\hline
Spatial-RNN\cite{wu2018deep} & M-Net/D-Net & - & 88.4 & - \\ 
\hline
BCN\cite{Dubey_2018_ECCV} & ResNet-50 & 87.7 & 90.3 & 94.3 \\
\hline
ACNet\cite{ji2020attention} & ResNet-50 & 88.1 & 92.4 & 94.6 \\
\hline
DCL\cite{chen2019destruction} & ResNet-50 & 87.8 & 93.0 & 94.5 \\
\hline
DF-GMM\cite{WangWYLLL20} & ResNet-50 & 88.8 & 93.8 & 94.8 \\
\hline
a-pooling\cite{SimonRDD20} & ResNet-50  &  86.5 & - & 91.6 \\
\hline
MA-CNN\cite{ZhengFML17}& VGG-19 &  86.5 & 89.9 & - \\
\hline
NTS\cite{YangLWHGW18} & ResNet-50 &  87.5 & 91.4 & 93.9 \\
\hline
API-net\cite{ZhuangW020} &ResNet-50  & 87.7 & 93.0 & 94.8 \\
\hline
GCL\cite{WangWLDL20} & ResNet-50 & 88.3 & 93.2 & 94.0 \\
\hline
MGE\cite{zhang2019learning} & ResNet-50 & 88.5 & - & 93.9  \\
\hline
CS-Parts\cite{korsch2019classification} & ResNet-50  &  89.5 & - & 92.5 \\
\hline
Inceptin-v3\cite{cui2018large} & Inception-v3   & 89.6 & 90.7 & 93.5 \\
\hline
PMG\cite{du2020fine} & ResNet-50  & 89.6 & 93.4 & 95.1  \\
\hline
Mix+\cite{abs-2004-02684} & ResNet-50  & 90.2 & 92.0 & 94.9 \\
\hline
StackedLSTM\cite{GeLY19}& GoogleNet &   90.4 & - & -\\
\hline
\hline
Our CCFR w$\backslash$o re-ranking & ResNet-50 & 90.7 & 93.0 & 95.37 \\
\hline
Our CCFR & ResNet-50  & \textbf{91.1} & \textbf{94.1} & \textbf{95.49} \\
\hline
\end{tabular}
\end{center}
\caption{Comparison of different methods on CUB-200-0211, FGVC Aircraft and Stanford Cars.}
\end{table*}

\section{Experiments \& Analysis}
\subsection{Datasets}
We conduct comprehensive experiments to evaluate our proposed CCFR algorithm on Caltech-UCSD Birds (CUB-200-2011\cite{wah2011caltech}), FGVC Aircraft\cite{maji2013fine} and Stanford Cars\cite{krause20133d} which are popular benchmarks for fine-grained category classification. The details of these three datasets are shown in Table 1. Note that we only use the image-level labels in our all experiments.

\noindent
\textbf{Caltech-UCSD Birds.}
CUB-200-2011 is the most widely used dataset with 200 wild bird species. It contains 11,788 images spanning 200 sub-categories. The ratio of train data and test data is roughly 1:1 and each species has only 30 images for training.

\noindent
\textbf{FGVC Aircraft.}
FGVC Aircraft dataset consists of 10,000 images with 100 categories. The ratio of train data and test data is roughly 2:1. The dataset is organized in a four-level hierarchy, from finer to coarser: Model, Variant, Family, Manufacturer. Most images are airplanes.

\noindent
\textbf{Stanford Cars.}
Stanford Cars dataset consists of 16,185 images over 196 categories. The data is divided into 8,144 training images and 8,041 testing images, and each category has been divided roughly into a 50-50 split. Categories are typically at the level of Make, Model, Year (e.g. 2012 BMW M3 coupe).

\subsection{Implementation Details}
In all our implementations, ResNet-50 is chosen as the feature feature extractor. We preprocess each image to size 448$\times$448, and select 2 local regions for each scale (there are two scales), and set the NMS threshold to 0.25. We adopt Momentum SGD with an initial learning rate 0.001 and multiply it by 0.1 for each 30 epochs, and set the weight decay to 1e-4, and set the batch-size to 16. We first utilize the ranking loss\cite{YangLWHGW18} to train the FPN for satisfactory detection performance, and then use the Softmax loss and Triplet loss to train our model. For re-ranking in Equation\ref{con:eq8}, we set the $topn$ as 5, and set $\alpha$ to 0 and $\beta$ to 1.0 which means only using the searching similarity when the Softmax probability is low. The threshold $T_{sf}$ is set differently according to the roughly average Top1 Softmax probabilities on different datasets (hear we set 0.5 on CUB-200-2011, and 0.7 on Cars and Airs). More analysis about these re-ranking parameters is given in the ablation experiments.

\subsection{Quantitative Analysis}
\begin{figure}[t]
\begin{center}
\includegraphics[width=1.0\linewidth]{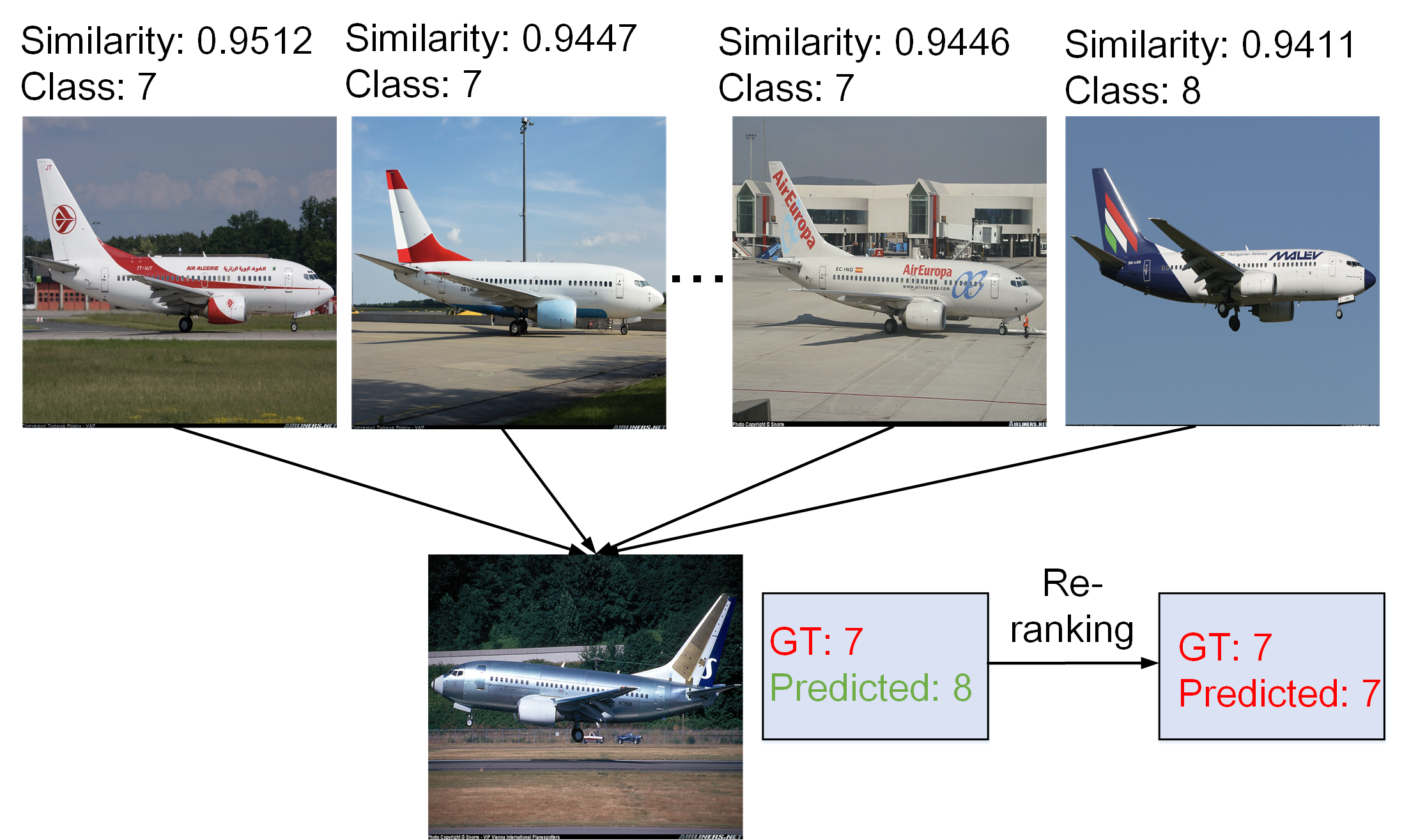}
\end{center}
   \caption{Rectifying a misclassified image by re-ranking. The ground-truth of the testing image is class $7$, and it is misclassified into class $8$ by a low classification probability. The $topm$ images with the highest similarity scores are retrieved from the database, and an new re-ranked score is calculated for each of the $topn$ classes according to Equation\ref{con:eq8}. Finally, the class $7$ with the biggest re-ranked score is taken as the recognized class.}
\label{re_ranking}
\end{figure}

We compare our CCFR with a number of recent works on three widely used benchmarks in Table 2. From the results, we find that our method outperforms all previous methods. To conduct fair comparisons with existing models, we use ResNet-50 as the feature extractor which is commonly used. Given that object/part annotations are labor-intensive or infeasible in practice, we train our CCFR with only image-level labels.


The third column of Table 2 shows the comparison results on CUB-200-2011. It shows that our CCFR achieves the highest Top1 accuracy and increases the accuracy by 0.7\%. Compared to NTS\cite{YangLWHGW18} which tries to obtain informative local regions with a ranking loss, we utilize the triplet loss to select more discriminative local regions, leading to better attention on discriminative areas such as bird’s head, wings and feet as shown in Fig~\ref{nts2ours}. Comparing with StackLSTM\cite{GeLY19} which is the best result in CUB-200-2011 up to now, we achieve a 0.7\% improvement. Besides, it is worth noting that StackLSTM is a very messy multi-stage training model that hampers the availability in practical use, while our proposed CCFR is easier to implement.

The fourth and fifth columns show the comparison results on Stanford Cars and FGVC Aircraft respectively. Our CCFR achieves new state-of-the-art results with 94.1\% Top1 accuracy on FGVC Aircraft and 95.49\% Top1 accuracy on Stanford Cars. In the second last row of Table 2, we show that our CCFR w$\backslash$o re-ranking also achieves satisfactory results which benefit from the elaborately designed architecture and effective training loss (i.e. the triplet loss for local regions, and the Multi-level loss for pretraining). 

\renewcommand{\arraystretch}{0.9}
\begin{table*}
\newcommand{\tabincell}[2]{\begin{tabular}{@{}#1@{}}#2\end{tabular}}
\begin{center}
\begin{tabular}{|l|c|c|c|c|c|} 
\hline
Model  & \tabincell{c}{Multi-level \\ loss} &  \tabincell{c}{Scale-separated \\ NMS} & \tabincell{c}{Fusion \\ Network} & Re-ranking & CUB Acc.(\%)  \\
\hline
ResNet-50  & &  & & & 84.5  \\
\hline 
ResNet-50 & \checkmark &  & & & 85.2 \\
\hline 
ResNet-50 + local region & \checkmark & & & & 90.3 \\ 
\hline
ResNet-50 + local region & \checkmark & \checkmark & & & 90.4 \\
\hline
ResNet-50 + local region & \checkmark & \checkmark & \checkmark & & 90.7 \\
\hline
CCFR & \checkmark  & \checkmark&   \checkmark &  \checkmark & 91.1 \\
\hline
\end{tabular}
\end{center}
\caption{
Ablation experiments for investigating the influence of different components about our method on CUB. The columns represent the different components (Multi-level loss, Scale-separated NMS, Fusion Network and Re-ranking). The rows represent the different models, where "ResNet-50 + local region" means that local regions are utilized in addition to the global features, and CCFR is our proposed method.
}
\end{table*}
Also, from the last two rows, it is clear that re-ranking can bring further improvement for the Top1 accuracy (especially when the Top1 softmax probability is relatively low, which usually means the model hesitates among the $topn$ classes). By re-ranking the $topn$ classes with the statistic of the $topm$ ($topm$=50) most similar retrievals from the database (usually constructed with the training samples), we probably rectify the uncertain Top1 class and rank the more confident class to Top1, and thus increase the Top1 accuracy (see Equation \ref{con:eq8}).
Fig~\ref{re_ranking} shows the misclassified image is rectified by re-ranking. Specifically, the ground-truth of the testing image is class $7$ which is misclassified into class $8$ by a low Softmax probability. Then we select the $topm$ images with the highest similarity scores from the database, and get an new score for each of the $topn$ classes according to our formulation, and finally take the class (class $7$) with the biggest new score as the recognized class.

\subsection{Ablation Experiments}
To investigate the influence of different components of our CCFR architecture, we conduct ablation studies and report the results. 

\textbf{Influence of Multi-level Loss.}
We investigate the influence of our Multi-level loss through experiments with standard Softmax loss. As shown in the second and third rows of Table 3, by applying our designed Multi-level loss on ResNet-50, the Top1 accuracy is improved by 0.7\%, which demonstrates its effectiveness. 
In this work, we utilize the Multi-level loss to pretrain the ResNet-50 as initialization of the backbone of our CCFR for better performance. 

\begin{table}
\begin{center}
\begin{tabular}{|l|c|c|c|} 
\hline
Method & CUB  & Airs & Cars  \\
\hline
Retrieval  &  88.8 & 93.4 & 91.54 \\
\hline 
Classification & 90.7 & 93.1 & 95.37 \\
\hline 
CCFR & 91.1 & 94.1 & 95.49 \\
\hline 
\end{tabular}
\end{center}
\caption{The comparison of the accuracy among Retrieval, Classification and CCFR on different datasets.}
\end{table}
\textbf{Influence of Scale-separated NMS.}
As shown in the fourth and fifth rows of Table 3, by applying the Scale-separated NMS, the performance improves from 90.3\% to 90.4\%, where the fourth row conducts the conventional NMS on all boxes collected from different scales, and the fifth row adopts the scale-separated NMS to reduce the mutual influence among different scale boxes.
The results verify the effectiveness of the Scale-separated NMS.

\textbf{Influence of Fusion Network.}
From the comparison between the fifth and the sixth rows in Table 3, we can find that the Top1 accuracy is improved from 90.4\% to 90.7\%, where 
the sixth row adopts Fusion Network (describe in section 3.2) to fuse the local region features obtaining from scale-separated NMS, while the fifth row directly concatenates these features after the global average pooling.

\begin{figure}[t]
\begin{center}
\includegraphics[width=1.0\linewidth]{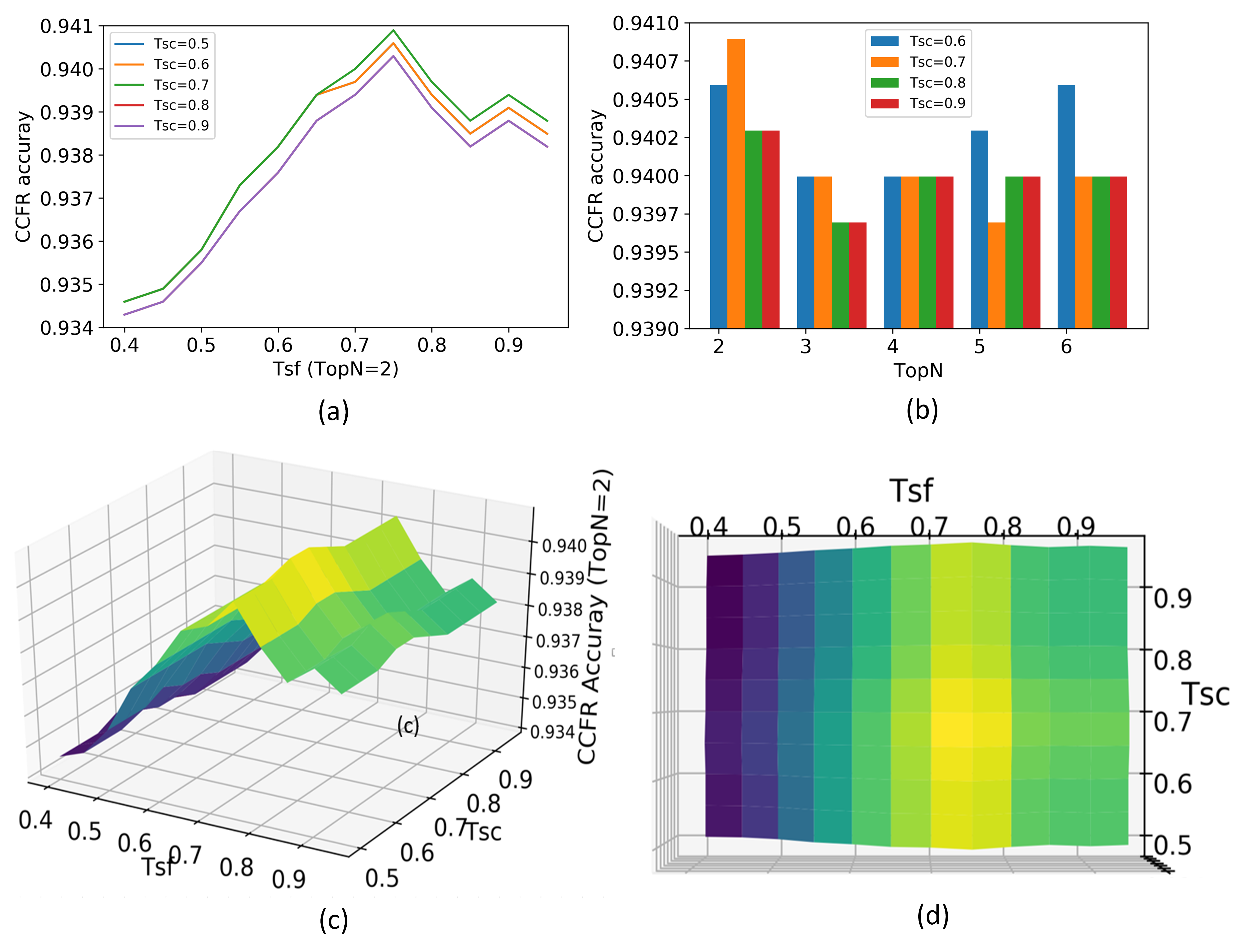}
\end{center}
   \caption{Ablation experiments for the main re-ranking parameters on FGVC Aircraft. Sub-figure (a) shows the influence of different $T_{sf}$ and $T_{sc}$ values ($topn$ is fixed to 2). Sub-figure (b) shows the impact of different $topn$ and $T_{sc}$ values while fixing $T_{sf}$ to 0.75.
   Sub-figure (c) and (d) visualize the surface over $T_{sc}$ and $T_{sf}$ respectively from different views, where warm color (yellow) means high accuracy and cold color (blue) represents low accuracy.
   }
\label{re_ranking_param}
\end{figure}

\textbf{Influence of Re-ranking.}
\begin{figure*}[t]
\begin{center}
\includegraphics[width=0.9\linewidth]{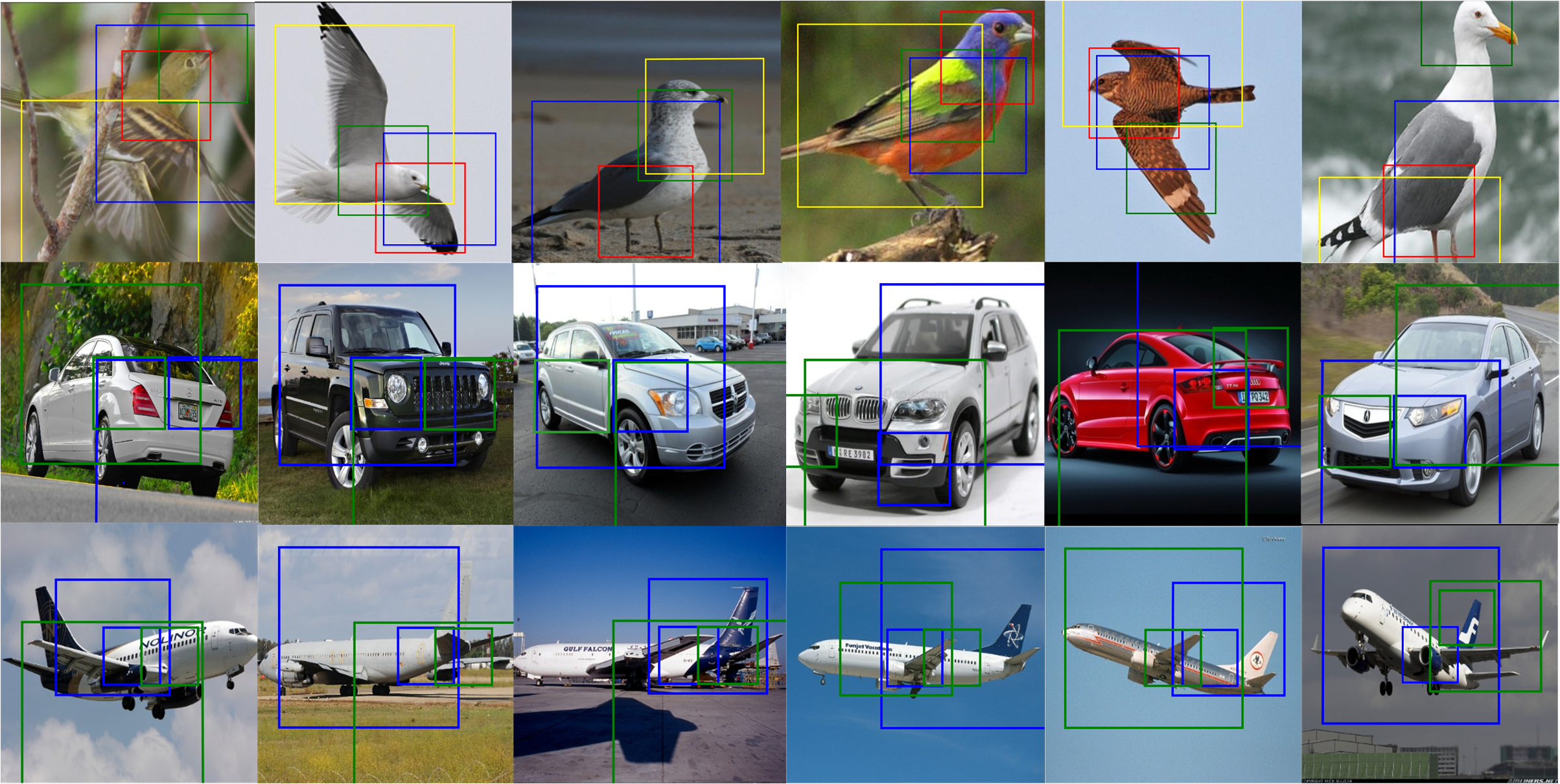}
\end{center}
   \caption{Illustration of the Top4 selected boxes by our model on three benchmark datasets. The bird's head and wings are deemed as the most crucial areas for birds, and for the cars are the light and head/tail, and for the airs are the turbine and tail.}
\label{visualize}
\end{figure*}
Since our CCFR is a retrieval-based model, we also compare the results obtained only by retrieval in Table 4. Specifically, we first compute the cosine similarity between the testing image and all the training images in database, and then use the Top1 image label as the final result. We can find that classification-based results are much better than
retrieval-based results on CUB-200-2011 and Stanford Cars (competitive on the FGVC Aircraft). And the last row in Table 4 verifies the idea that the combination of the retrieval and the classification can bring performance improvement. To further investigate the influence of the main re-ranking parameters, i.e. $T_{sf}$, $T_{sc}$ and $topn$ as described in Equation\ref{con:eq8}, more ablation experiments are conducted on the FGVC Aircraft dataset. Specifically, we investigate the change of the CCFR accuracy by ranging $topn$ from 2 to 6, $T_{sf}$ from 0.4 to 0.95, and $T_{sc}$ from 0.5 to 0.95 (Note that $topm$ is completely determined by $T_{sc}$, $\alpha$ and $\beta$ are simply set to 0 and 1.0). All the results as shown in Fig~\ref{re_ranking_param}. From Fig~\ref{re_ranking_param} (a), we can find that for different $T_{sc}$, the curves have the same trend and CCFR accuracy achieves the highest value when $T_{sc}$ is 0.7 and $T_{sf}$ is 0.75 (here $topn$ is fixed to 2, and a consistent phenomenon can be observed for other $topn$ values). In Fig~\ref{re_ranking_param} (b), we compare the effect of different $T_{sc}$ and $topn$ values while fixing $T_{sf}$ to 0.75 (the optimal value in (a)), and we can find the accuracy achieves best when $T_{sc}$ is 0.7 and $topn$ is 2 (it seems that most misclassified cases occur between the fist and second place ranked by the Softmax probabilities). With $topn$ fixed to 2, Fig~\ref{re_ranking_param} (c) (d) visualize the surface of the CCFR accuracy over $T_{sc}$ and $T_{sf}$ respectively from different views, where warm color (yellow) represents high accuracy and cold color  (blue) represents low accuracy. It can be clearly see that, $T_{sf}$ has the greatest influence on the re-ranking effect (which is reasonable), and the accuracy is insensitive to $T_{sc}$ (which means a wide range of $T_{sc}$ could achieve similar accuracy).
 



\subsection{Qualitative Analysis}
We draw the selected boxes in Fig~\ref{visualize} to better analyze the regions where our model focuses on. Two boxes with the highest confidence score are shown for each scale. 
It can be seen that, except for the region containing the whole object, our model deems head and wing as the most crucial areas for birds (the top two rows), and the light and head/tail for cars, and the turbine and tail for airs. All of these are consistent with human intuition.

\section{Conclusion}
In this work, we propose a Coarse Classification and Fine Re-ranking framework for the FGVC problem and get state-of-the-art results on three common datasets, where local region enhanced features are generated to re-rank the $topn$ classification results predicted by the semantic global features. We design an effective structure to integrate the local and global features, and use the triplet loss to discover more discriminative regions for distinguishing hard samples (supervised with only image-level labels). And to learn more effective semantic global features, we design a Multi-level loss acted on an auto-clustered hierarchical category structure for the backbone network pretraining. Some ablation experiments are provided for further analysis, which makes the work more convinced and solid.


{\small
\bibliographystyle{ieee_fullname}
\bibliography{egbib}
}

\end{document}